\documentclass{article}
\usepackage{ijcai21}

\usepackage{times}
\usepackage{soul}
\usepackage{color}
\usepackage{url}
\usepackage[hidelinks]{hyperref}
\usepackage[utf8]{inputenc}
\usepackage[small]{caption}
\usepackage{graphicx}
\usepackage{amsmath}
\usepackage{amsthm}
\usepackage{booktabs}
\usepackage{algorithm}
\usepackage{algorithmic}
\title{ Exploring Instance Relations for Unsupervised Feature Embedding }

\author{
Yifei Zhang$^{1,2}$
\and
Yu Zhou$^1$\footnote{Corresponding author}\and
Weiping Wang$^1$
\affiliations
$^1$Institute of Information Engineering, Chinese Academy of Sciences\\
$^2$School of Cyber Security, University of Chinese Academy of Sciences\\
\emails
\{zhangyifei0115, zhouyu, wangweiping\}@iie.ac.cn
}

\begin{document}

\maketitle

\begin{abstract}
Despite the great progress achieved in unsupervised feature embedding, existing contrastive learning methods typically pursue view-invariant representations through attracting positive sample pairs and repelling negative sample pairs in the embedding space, while neglecting to systematically explore instance relations.
In this paper, we explore instance relations including intra-instance multi-view relation and inter-instance interpolation relation for unsupervised feature embedding.
Specifically, we embed intra-instance multi-view relation by aligning the distribution of the distance between an  instance's different augmented samples and negative samples. 
We explore inter-instance interpolation relation by transferring the ratio of information for image sample interpolation from pixel space to feature embedding space.
The proposed approach, referred to as EIR, is simple-yet-effective and can be easily inserted into existing view-invariant contrastive learning based methods.
Experiments conducted on public benchmarks for image classification and retrieval report state-of-the-art or comparable performance.
%
%$\footnote{$^*$Corresponding authors\\}$
%
Our code will be available at \url{https://github.com/zhangyifei01/EIR}.

\end{abstract}

\section{Introduction}
Unsupervised feature embedding methods have attracted increased attention in recent years.
As a mainstream method for unsupervised feature embedding, contrastive learning method focuses on learning an embedding function by attracting positive pairs and repelling negative pairs. However, it is challenging to discover positive and negative samples under unsupervised scenarios. 
%%

%%%%%%%%%%%%%%%%%%%%%%%%%%
\begin{figure}
\centering
% width=1.0\columnwidth
\includegraphics[width=0.87 \columnwidth]{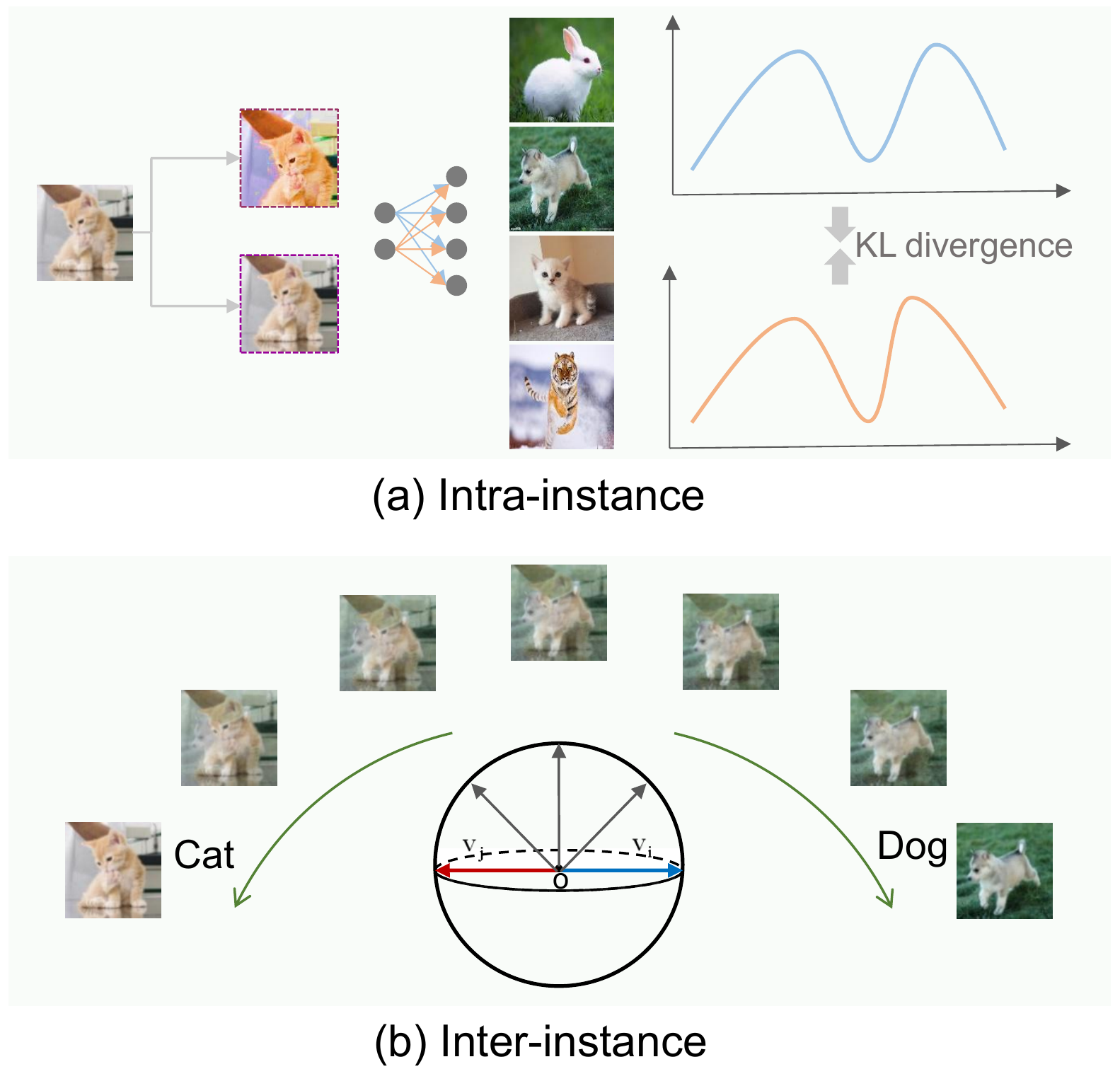}
\caption[]{ Illustrations of the instance relations including both intra-instance multi-view relation (a) and inter-instance interpolation relation in an unit sphere embedding space (b).
}
\label{fig:fig1}
\end{figure}

To discover reliable positive and negative samples on unlabeled datasets, instance discrimination based unsupervised methods have achieved promising results on unsupervised feature embedding. Instance Recognition (IR)~\cite{Wu_2018_IR} proves that non-parametric instance-wise classification can capture apparent visual similarity. Invariant and Spreading Instance Feature (ISIF)~\cite{Ye_2019_IS} 
exploits data augmentation invariant and instance spreading property for unsupervised learning. Momentum Contrast (MoCo)~\cite{He_2020_Moco} build a dynamic dictionary with a queue and a moving-averaged encoder. After that, SimCLR~\cite{Chen_2020_SimCLR} simplifies these instance discrimination based contrastive learning algorithms without requiring specialized architectures.
Compared with instance discrimination based unsupervised methods, supervised learning methods absorb more information from the different instances with same semantic categories.
While instance discrimination based methods have shown their effectiveness on several benchmarks, they severely rely on single instance discrimination and data augmentation, and ignore the relations of instances, Figure~\ref{fig:fig1}.
Augmentation Invariant and Spreading Instance Feature (aISIF)~\cite{Ye_2020_Embedding} improves ISIF by two feature-level augmentation strategies including negative augmentation with interpolation and positive augmentation with extrapolation. However, the instance relations are still not systematically explored, which makes the learned feature embedding less discriminative.
%

%%%%%%%%%%%%%%%%%%%%%%%%%%%%%%%%%%%%%%%%%%%%%
\begin{figure*}
\centering
% width=1.0\columnwidth
\includegraphics[width=1.9\columnwidth]{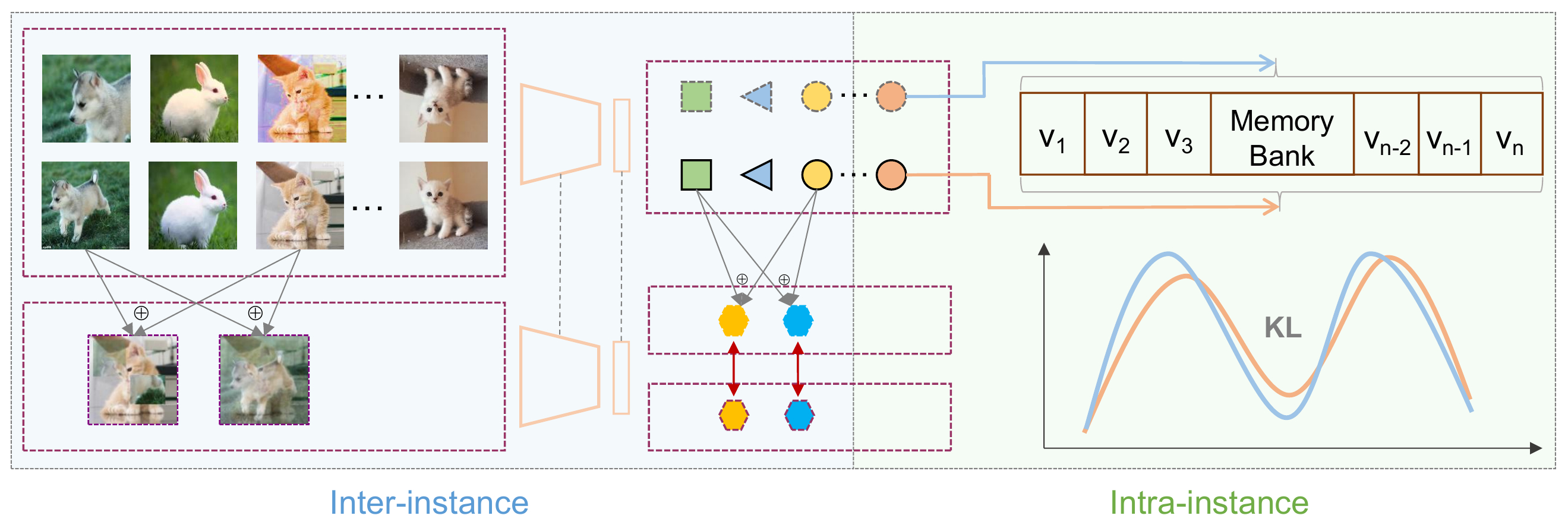}
\caption[]{Overview of EIR, which includes intra-instance multi-view relation (left) and inter-instance interpolation relation (right). Different shapes represent different semantic categories, and different colors represent different instances.
}
\label{fig:fig2}
\end{figure*}

In this paper, we present exploring instance relations (EIR) including both intra-instance multi-view relation and inter-instance interpolation relation for unsupervised feature embedding, Figure~\ref{fig:fig2}.
Specifically, for exploring more sophisticated intra-instance relations, we align two distributions with KL divergence, each of which is the distance distribution
of an augmented sample with respect to all samples in the training set, Figure~\ref{fig:fig1}(a). 
Benefiting from current data augmentation strategies such as mixup~\cite{mixup_2018_iclr} and cutmix~\cite{cutmix_2019_iccv}, we explore inter-instance interpolation relation by transferring the ratio of information for image sample interpolation from pixel space to feature space, Figure~\ref{fig:fig1}(b).
By simultaneously incorporating these two relations, our approach improves the capability of discrimination in the embedding space.
Our approach is simple-yet-effective and can be easily inserted into existing view-invariant contrastive learning based methods.

We present extensive evaluations of our approach on various datasets and tasks to demonstrate its effectiveness. In particular, our method achieves the accuracy of 89.2\% on CIFAR-10 in the kNN setting, which is the state-of-the-art performance to the best of our knowledge. 

%or comparable performance over other unsupervised methods with kNN top-1 accuracy 87.6\% on CIFAR-10, 78.2\% on STL-10 and 66.5\% on ImageNet-100.%

The contributions of this work are summarized as follows: 
\begin{enumerate}
		\item We propose EIR to systematically explore previous overlooked instance relations for unsupervised feature embedding, which is a novel attempt to break through the limit of instance discrimination.   
		\item We explore the intra-instance multi-view relation by aligning two augmented samples' distance distribution, which measures the relations of the augmented samples with respect to all samples in the training set.
		
		\item We explore inter-instance interpolation relation by transferring the ratio of information for image sample interpolation from pixel space to feature space.
		
		\item We conduct experiments for image classification and retrieval on CIFAR-10, STL-10, ImageNet-100, CUB-200 and Car-196 datasets. EIR achieves state-of-the-art or comparable performance.
		
% 		\item Extensive evaluations are conducted on several public benchmarks, and demonstrates the effectiveness of EIR, the intra-instance multi-view relation and the inter-instance interpolation relation.
\end{enumerate}

% The rest of the paper is organized as follows. We first review related works in Section 2, then the details of the proposed method are explained in Section 3. In Section 4, the implementation and results of the experiments are provided and analyzed. Finally, we conclude our works in Section 5.

\section{Related Works}
%
% Unsupervised deep learning methods have drawn increased attention in recent years. 
There are two related concepts in unsupervised deep learning methods: unsupervised representation learning and unsupervised embedding learning.  
%区别到底是啥呢？

{\bf Unsupervised Representation Learning.}
Unsupervised representation learning aims at learning `immediate' $conv$ features, then the features can be used as the initialization for downstream tasks such as image classification and object detection. Some works focus on designing pretext tasks by hiding some information such as context prediction~\cite{Doersch_2015_Context}, jigsaw puzzle~\cite{Noroozi_2016_Jigsaw}, colorization~\cite{Guatav_2016_Color}, split-brain~\cite{Zhang_2017_splitbrain} and rotations prediction~\cite{Gidaris_2018_Rotnet}. 

%%%
Unsupervised contrastive learning based methods can be roughly categorized into two categories: instance-based methods and group-based methods.
For instance-based methods, IR~\cite{Wu_2018_IR} treats each sample as an independent class. Contrastive Multiview Coding(CMC)~\cite{Tian_2019_CMC} proposes to maximize the mutual information between different views and scales to any number of views. MoCo~\cite{He_2020_Moco} builds a dynamic dictionary with a queue and a moving-averaged encoder. After that, SimCLR~\cite{Chen_2020_SimCLR} simplifies these instance discrimination based contrastive learning algorithms without requiring specialized architectures or memory bank.
For group-based methods, \cite{Yang_2016_CVPR} and DeepCluster (DC)~\cite{Caron_2018_DC} joint clustering and feature learning by iteratively grouping features and using the assignments as pseudo labels to train feature representations. Anchor Neighbourhood Discovery (AND) \cite{Huang_2019_AND} adopts a divide-and-conquer method to find local neighbours. Local Aggregation~\cite{Zhuang_2019_LA} trains an embedding function to maximize a metric of local aggregation. Online DeepCluster~\cite{Zhan_2020_ODC} performs clustering and network updating simultaneously rather than alternatively.
% PCP~\cite{Zhang_2020_PCP} propose progressive clustering and cluster purification mechanism to improve class consistency.
%
% Existing group-based methods remain challenged by the significant number of noise samples during clustering or neighbourhood discovering, which deteriorate the performance of representation models.

{\bf Unsupervised Embedding Learning.}
Unsupervised embedding learning focuses on learning a low-dimension feature embedding by minimizing the distance of positive pairs and maximizing the difference of negative pairs, and the learned $conv$ features also can be used for initialization. 
Mining on Manifolds~\cite{mom_2018_metric} proposes an unsupervised method to mine hard positive and negative samples based on manifold space. ISIF~\cite{Ye_2019_IS} proposes to utilize the instance-wise supervision to approximate the positive concentration and negative separation properties, and aISIF~\cite{Ye_2020_Embedding} improves ISIF by two feature-level augmentation strategies including negative augmentation with interpolation and positive augmentation with extrapolation. PSLR~\cite{Ye_2020_PSLR} incorporates adaptive softmax embedding in graph latent space. Current contrastive based unsupervised representation learning methods, $i.e.$ IR~\cite{Wu_2018_IR}, MoCo~\cite{He_2020_Moco}, can also be applied for unsupervised embedding learning. 
%As demonstrated in~\cite{Ye_2020_Embedding}. contrastive based unsupervised representation learning methods usually perform bad on the challenging unseen testing categories, where training and testing set do not contain the same categories.%

\section{Methodology}

%%%%%%%%%%%%%%%%%%%%%%%%%%%%%%%%%%%%%%%%%%%%%%%%%%%%%%%%
\begin{figure}
\centering
% width=1.0\columnwidth
\includegraphics[width=1\columnwidth]{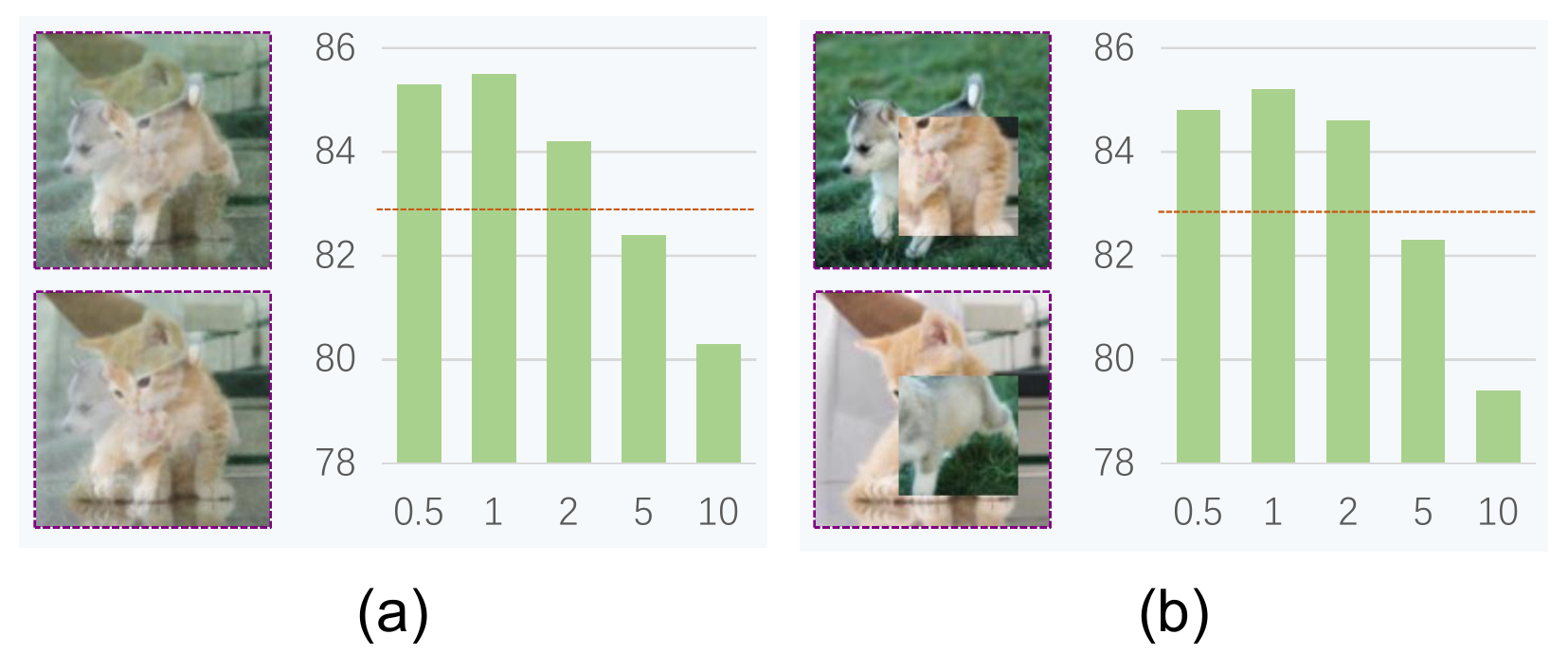}
\caption[]{ Illustrations of the instance interpolation operations and its performance under different coefficients. Dashed line denotes the accuracy for baseline method. (a) Mixup~\cite{mixup_2018_iclr}; (b) Cutmix~\cite{cutmix_2019_iccv}.
}
\label{fig:fig3}
\end{figure}

\subsection{Preliminaries}
Given an imagery data set $X = {\{x_{1}, x_{2}, x_{3}, \cdots, x_{N}\}}$ without any manual annotations, the learned embedding model with parameters $\theta$ maps an input image $x_{i}$ to a $D$-dimension embedding in feature space $\mathcal{V}(t)$ at $t$-th epoch, as $v_{i}(t) = f_{\theta_{t}}(x_{i})$. The objective is to learn an embedding feature space where embedding for positive pairs are attracted while negative pairs are repelled.  Details will be illustrated as follows.

IR~\cite{Wu_2018_IR} proves that CNNs can capture apparent similarity and learn class discriminative feature representation with solely instance-level supervision. The probability of input $x$ being recognized as $i$-th example is
\begin{equation}
    P(i|v) = \frac{exp(v_{i}^{T}v/\tau)}{\sum_{j=1}^{N}exp(v_{j}^{T}v/\tau)} ,
    \label{p-ir}
\end{equation}
where $v = f_{\theta}(x)$, $N$ is the size of whole dataset and $\tau$ is a temperature parameter. Objective function is defined as $L = -\sum_{i=1}^{N} log P(i|f_{\theta}(x_{i}))$. IR maintains a memory bank for storing the global features of dataset, $v_i$ and $v_j$ in Eq.(\ref{p-ir}) denote the $i$-th and $j$-th features in memory bank. Memory bank is initialized as unit random vectors and evolves during training iteration by $v_{i}(t) = \ell_{2}((1 - m) \times v_{i}(t) + m\times v_{i}(t-1))$. 
% The feature stored in memory bank evolves by moving-averaged features with random data augmentations and different training stage of CNNs, which can be regarded as 'prototype' of each instance.

%%%%%%%%%%%%%%%%%%%%%%%
Based on the analysis of IR, we modify instance recognition by applying multi-view data augmentation, which termed IRaug. The probability of an input $x$ and its another view $\hat{x}$ being recognized as $i$-th example is
\begin{equation}
    P(i|v,\hat{v}) = \frac{exp(v_{i}^{T}v/\tau)}{\sum_{j=1}^{N}exp(v_{j}^{T}v/\tau)} + \frac{exp(v_{i}^{T}\hat{v}/\tau)}{\sum_{j=1}^{N}exp(v_{j}^{T}\hat{v}/\tau)} .
    \label{p-iraug}
\end{equation}
IRaug forces $v$ and $\hat{v}$ to be close to the corresponding feature $v_i$ in memory bank, is not just implementing IR twice.  More specifically, it forces $v$ and $\hat{v}$ to be close to each other implicitly. We optimize the objective function by 
\begin{equation}
    L_{IRaug} = -\sum_{i=1}^{N} log P(i|f_{\theta}(x_{i}), f_{\theta}(\hat{x}_{i})).
    \label{l-iraug}
\end{equation}
We update memory bank with either of $v$ or $\hat{v}$. IRaug will serve as our baseline in the following experiments.
% There are two results shows the difference between IR and IRaug: a) Comparing IRaug training $k$ rounds and IR with $2k$ rounds until convergence, IRaug superior than IR with significant margins; b) IRaug is slightly slower than IR in training same round, but faster a lot than IR with two times rounds.

\subsection{Intra-instance Multi-view Relation}
%%%%%%%%%%%%%%%%%%%%%%%%%%%%%%%%%
From the perspective of ISIF~\cite{Ye_2019_IS} and MoCo~\cite{He_2020_Moco}, view-invariant contrastive loss focuses on attracting different views of the same instance while repelling the different instances. However, more sophisticated intra-instance relations are not fully explored.
% View-invariant loss attracts different views of same instance explicitly, which is defined as
% \begin{equation}
%     L_{invariant} = - \sum_{i=1}^{n} log \frac{exp(v_{i}^{T}\hat{v}_{i}/\tau)}{\sum_{j=1}^{n_{b}}exp(v_{j}^{T} \hat{v}_{i}/\ tau)}.
%     \label{p-view-invariant}
% \end{equation}
% Considering the baseline method, we can form the loss with coefficient $\lambda_{0}$ by
% \begin{equation}
%     Loss_{0} = L_{IRaug} + \lambda_{0} L_{invariant}.
%     \label{l-add-invariant}
% \end{equation}
%View-invariant loss function is a strict limitation.%

One such relation is that the current instance's two augmented samples should have similar distances with each sample in the training set. As illustrated in the right part of Figure~\ref{fig:fig2}. Concretely, other than only pushing negative samples far away, two augmented samples from the same instance should push the same negative sample to similar distances. Formally, for the two augmented samples from the same instance, two corresponding distance distributions can be obtained: $D$ and $\hat{D}$,
\begin{equation}
    D = \{d_{i}=v^{T}v_{i} | i=1,2,3,...,N\},
\end{equation}
 and 
\begin{equation}
     \hat{D} = \{\hat{d}_{i}=\hat{v}^{T}v_{i} | i=1,2,3,...,N\}.
\end{equation}
%Our hypothesis is that good feature embedding for instance should be maintaining a relation that the distribution of the distance between an identical instance of different views and negative  samples can be aligned. %
These two distributions should be aligned. We define this relation as intra-instance multi-view relation, which can be modeled by the Kullback–Leibler divergence~\cite{KL} of distributions $D$ and $\hat{D}$. By this way, the objective function is formed by

%For each view, the distance between this view and all features stored in memory bank can be calculated within the mini-batch size. Different views of identical instance should share similar distance distribution. We maintain the multi-view intra-instance relation by aligning the distance distribution of different views. It allows our model to capture view-invariant information and embedding space position information by maintaining the multi-view intra-instance relation with the usage of negative samples.%
%Formally, 

 \begin{equation}
     L_{intra} = \sum_{p=1}^{N} KL(D(p)|| \hat{D}(p)),
     \label{l-intra}
 \end{equation}
where $D(p)$ denotes the distant distribution of current instance, and $\hat{D}(p)$ is that of another view. 
% %Combining baseline with intra-instance multi-view relation objective function, the loss with coefficient $\lambda_{1}$ is represented by%
% %\begin{equation}
%     Loss_{1} = L_{IRaug} + \lambda_{1} L_{intra}.
%     \label{l-add-intra}
% \end{equation}
%

\subsection{Inter-instance Interpolation Relation}
% some works explore instance interpolation as negative samples. %
To address inter-instance relation, aISIF~\cite{Ye_2020_Embedding} explores feature-level interpolation as negative samples by $v_{p-} = \ell_{2}(\alpha v_{i} + (1-\alpha) v_{k})$, which consistently improves the performance. Inspired by this work, we move one more step to explore the inter-instance relation from pixel level to feature level.

Our hypothesis is that better features can be learned if the consistency of the pixel-level interpolation and the feature-level addition is 
maintained. As illustrated in the left part of Figure~\ref{fig:fig2}, if an image is composed of two images with some specific ratios by interpolation, its feature should be close to the addition of the two images' features with corresponding ratios.
%The similarity between instance pairs and their interpolated instance should be transferred in embedding feature space.%
Benefiting from current data augmentation strategies such as mixup~\cite{mixup_2018_iclr} and cutmix~\cite{cutmix_2019_iccv}, inter-instance interpolation can be implemented by mixup or cutmix. More general, we explore interpolation by information fusion with different ratios, Figure~\ref{fig:fig3} shows some possible interpolation alternatives and its performance.

\begin{figure}
\centering
% width=1.0\columnwidth
\includegraphics[width=0.9\columnwidth]{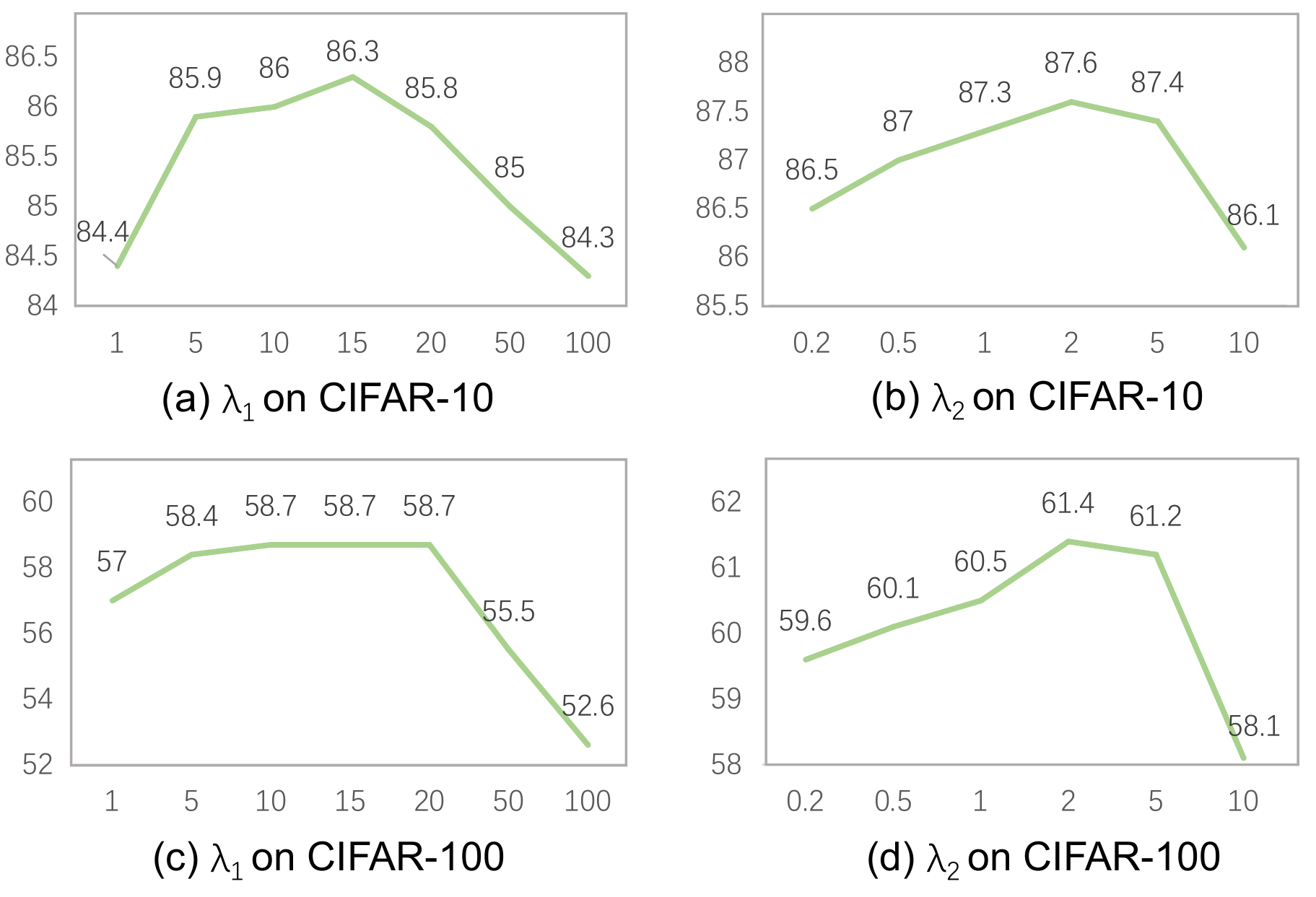}
\caption[]{ Effect of EIR with coefficient $\lambda_{1}$ and $\lambda_{2}$ on CIFAR-10 and CIFAR-100 datasets.
}
\label{fig:fig4}
\end{figure}

%---------------------------------------
% Table 1
\setlength{\tabcolsep}{4pt}
\begin{table}
\begin{center}
\caption{Effect of the components in our approach with $k$NN top-1 accuracy on CIFAR-10, $k=200$. }
\label{tab:table1}
\begin{tabular}{cccl}
\hline\noalign{\smallskip}
IRaug  &Intra-instance &Inter-instance  &Accuracy(\%)  \\
\noalign{\smallskip}
\hline
\noalign{\smallskip}
$\surd$  &- &-  &83.4  \\
      $\surd$  &$\surd$ &-   &86.3 \textcolor{red}{(+2.9)}  \\
      $\surd$  &- &$\surd$  &85.5 \textcolor{red}{(+2.1)} \\
      
      $\surd$  &$\surd$ &$\surd$   &87.6 \textcolor{red}{(+4.2)} \\
      
\hline
\end{tabular}
\end{center}
\end{table}
\setlength{\tabcolsep}{1.4pt}

For two random instances $x_{i}$ and $x_{j}$ in unlabelled dataset, we can extract partial information from $x_{i}$ with ratio $r$ and partial information from $x_{j}$ with ratio $(1-r)$ to form a new image sample. The feature embedding of interpolated sample is defined as
\begin{equation}
    v^{k} = f_{\theta}(r x_{i} \oplus (1-r) x_{j}),
\end{equation}
where $\oplus$ denotes interpolation operation, and $r$ denotes the ratio of information for interpolation. Meanwhile, the feature addition in embedding space is generated by 
\begin{equation}
    \Tilde{v}^{k} = \ell_{2}(r f_{\theta}(x_{i}) + (1-r) f_{\theta}(x_{j})),
\end{equation}
where $\ell_{2}$ denotes $\ell_{2}$ normalization.
By attracting these two features, the information ratio can be transferred from pixel space to feature space. The inter-instance interpolation relation loss is obtained by 
\begin{equation}
    L_{inter} = -log \sum_{k=1}^{N}\frac{exp((v^{k})^{T}\Tilde{v}^{k}/\tau)}{\sum_{j=1}^{N}exp((v^{j})^{T}\Tilde{v}^{k}/\tau)}.
\end{equation}

In summary, the overall loss function is a combination of the baseline loss $L_{IRaug}$, the intra-instance multi-view relation loss $L_{intra}$ and the inter-instance interpolation relation loss $L_{inter}$, which is formulated as
\begin{equation}
    Loss = L_{IRaug} + \lambda_{1} L_{intra} + \lambda_{2} L_{inter}.
    \label{l-all}
\end{equation}

%%%%%%%%%%%%%%%%%%%%%%%%%%%%%%%%%%%%%%%%%%%%%%%%%%%%%%%%
% \begin{figure}
% \centering
% % width=1.0\columnwidth
% \includegraphics[width=0.8\columnwidth]{fig4_3.pdf}
% \caption[]{ Effect of conventional contrastive loss with two augmentations on CIFAR-10 and CIFAR-100 dataset. Dashed line denotes the performance of baseline method IRaug.
% }
% \label{fig:fig4}
% \end{figure}
%%%%%%%%%%%%%%%%%%%%%%%%%%%%%%%%%%%%%%%%%%%%%%%%%%%%%%%%

\section{Experiments}

We conduct extensive experiments to evaluate the proposed method on downstream tasks including classification and retrieval. 
% More details are described in the supplementary material.

%--------------------------
% Table 2
\setlength{\tabcolsep}{10pt}
\begin{table}
\caption[]{Comparison of kNN top-1 accuracy with different $k$ on CIFAR-10 dataset. $R$ denotes training round (200 epochs for 1 round). Performance of other methods is copied from PSLR~\cite{Ye_2020_PSLR} and aISIF~\cite{Ye_2020_Embedding}.}
\begin{center}
\begin{tabular}{l|c|c|c}
\hline
\noalign{\smallskip}
Methods &k=5 &k=20 &k=200 \\
\noalign{\smallskip}
\hline
\hline
\noalign{\smallskip}

   Random  &32.4 &34.8 &33.4  \\
%   Super-posonly &- &- &$fail$ \\
%   Super-neg &91.1 &91.4 &91.5 \\
   
%   Super-pos &93.7 &93.8 &93.7 \\
   \hline
   \noalign{\smallskip}
%   DC$^*$ (1000) &- &- &- \\
%   Exemplar  &73.2 &74.0 &74.5 \\
  IR  &79.6 &80.5 &80.8 \\
%   IR(NCE) &79.4 &80.2 &80.4 \\
  DC(1000) &66.5 &67.4 &67.6  \\
%   DC$^*$(1000)  &79.7 &80.4 &80.1  \\
%   CPC &81.6 &82.1 &82.8 \\
  ISIF &82.4 &83.1 &83.6 \\
%   AET &77.6 &76.3 &78.2 \\
%   AVT &78.4 &78.5 &79.0 \\
   
%   CMC (Tian et al. 2019) &82.0 &82.6 &83.1 \\
  CMC &82.0 &82.6 &83.1 \\
%   AND(2$R$s) &82.7 &83.6 &84.2 \\
%   PCP &83.7 &84.7 &84.7 \\
   aISIF &86.4 &87.0 &86.6 \\
   PSLR &83.8 &84.7 &85.2 \\
   \hline\noalign{\smallskip}
   IRaug  &81.5 &83.4 & 83.4  \\
   Ours   &\bf{86.6} &\bf{87.5} &\bf{87.6} \\
   \hline \hline
   \noalign{\smallskip}
   AND(5$R$s) &84.8 &85.9 &86.3 \\
%   PCP(5$R$s) &86.1 &86.8 &87.3 \\
   PSLR$^+$(5$R$s) &87.4 &88.1 &88.4\\
   \hline \noalign{\smallskip}
%   Ours(1$R$)+coslr+proj   &- &- &\bf{-} \\
   Ours(2$R$s)   &\bf{87.7} &\bf{88.3} &\bf{88.6} \\
   Ours(5$R$s)   &\bf{88.2} &\bf{89.2} &\bf{89.1} \\

\hline
\end{tabular}

\label{tab:table2}
\end{center}
\end{table}
\setlength{\tabcolsep}{1.4pt}

\subsection{Experimental setting}
\textbf{Datasets.}
CIAFR-10/CIFAR-100~\cite{CIFAR} contains 50K training and 10K test images with 10/100 object classes, and all images are with size 32 $\times$ 32. STL-10~\cite{STL} contains 5K labeled images with 10 classes and 100K unlabeled images for training, and 8K images with 10 classes for testing, and all images are with size 96 $\times$ 96. During pre-training, the labels are deleted. ImageNet~\cite{ImageNet-IJCV15} subset ImageNet-100 contains 126,689 images with 100 classes for training, and 5K images for testing, and all images are resized to 224 $\times$ 224.

We split the 200 classes of CUB-200~\cite{CUB} into two parts. The first 100 classes with 5,864 images are used for training, and the last 100 classes with 5,924 images are used for testing, and all images are resized to 224 $\times$ 224. We then split the 196 classes of Car-196~\cite{Car} into two parts. The first 98 classes with 8,054 images are used for training, other 98 classes with 8,131 images are uased for testing, and all images are resized to 224 $\times$ 224.

\textbf{Hyper-parameters.}
We adopt ResNet18~\cite{ResNet} as the backbone, and initialize the learning rate as $0.03$ with the decreasing strategy that the rate is scaled down by $0.1$ at $120$-th epoch and $0.01$ at $160$-th epoch. We set momentum to $0.9$, weight decay to $0.0005$, batch size to $128$, temperature parameter $\tau$ to $0.1$, and the embedding feature space dimension to $128$. All experiments are implemented with 2 GeForce RTX 2080Ti GPUs. 
% Basic data augmentation operations include $RandomResizedCrop(scale=(0.2, 1.))$, $ColorJitter(0.4, 0.4, 0.4, 0.4)$, $RandomGrayscale(p=0.2)$ and $RandomHorizontalFlip()$.

\textbf{Evaluation Metrics.}
We use a single center-croped image for feature extraction in testing phase,
and use linear classification and $k$NN to evaluate features on classification task. 
% With linear classification, we implement classifier by designing a fully connected (FC) layer. We are aiming to optimize a FC layer by minimizing cross-entropy loss. With weighted $k$NN, for a test image $\hat{x}$, we calculate the cosine similarity $s_i$ = cos($v_i$, $\hat{v}$) for each of sample $x_i$ belonging to the train set, where $\hat{v}$ = $f_\theta$($\hat{x}$). The set of top-$k$ nearest neighbours denoted as $\mathcal{N}_k$ is then used to predict the class according to $\omega_c$ = $\sum_{i\in \mathcal{N}_k} \alpha_i \cdot 1 (c_i = c)$, where $\alpha_i = exp(s_i / \tau)$ and $\omega_c$ denotes the value of voting function that sample $\hat{x}$ belongs to class $c$. The result label of sample $\hat{x}$ is $c^*$ when $\omega_{c^*}$ = $\max$($\omega_0, \omega_1, \dots, \omega_{n-1}$), where $n$ denotes the number of categories in test set.
% R@K 
The retrieval performance is evaluated by the probability of correct matching  in the top-$k$ retrieval ranking list.
% To evaluate retrieval performance, cosine similarity is adopted for similarity measurement. For the query image from the test set, $R@K$ (\%) accuracy represent the probability of any correct matching with same category label occurs in the top-$k$ retrievaled ranking list. The final score is the average score for all test images.
%

\subsection{Ablation Study}
We take IRaug as our baseline method, then analyze the effect of our proposed module. In Table~\ref{tab:table1}, the baseline method achieves 83.4\% top-1 $k$NN ($k$=200) accuracy on CIFAR-10 dataset with ResNet18. The experimental results show that compared with the baseline, intra-instance multi-view relation improves the accuracy by 2.9\%, inter-instance interpolation relation improves the accuracy by 2.1\%, moreover, combining these two components can improve the accuracy by 4.2\%. 
% Table~\ref{tab:table1} shows the effectiveness of our proposed method.

%
% We then analysis the performance of conventional contrastive learning loss Eq.~\ref{l-add-intra} in Figure~\ref{fig:fig4}. More general, we conduct experiments on CIFAR-10 and CIFAR-100 for evaluating the effect of  coefficient $\lambda_{0}$. Baseline method achieves 83.4\% on CIFAR-10 and 55.6\% on CIFAR-100. By the time $\lambda_{0}$ equals to $0.5$, we achieve best performance, which improves baseline by 1.0\% and 1.4\% respectively, Figure~\ref{fig:fig4}.

% 
In Figure~\ref{fig:fig4} (a) and (c), we analyze the effect of the coefficient $\lambda_{1}$ in our loss function by incorporating baseline and intra-instance multi-view relation loss on CIFAR-10 and CIFAR-100 dataset. The experimental results show that with proper $\lambda_{1}$($\lambda_{1}$=15) our model achieves 86.3\% and 58.7\% top-1 accuracy on CIFAR-10 and CIFAR-100, respectively. 
% For the analysis the performance in Figure~\ref{fig:fig4} (a)(b) and Figure~\ref{fig:fig5} (a)(c), our proposed multi-view intra-instance relation shows more advantage over conventional view-invariant contrastive loss.
%
We then analyze the coefficient $\lambda_{2}$ of inter-instance interpolation loss. The effect of $\lambda_{2}$ is shown in Figure~\ref{fig:fig4} (b) and (d)  by fixing $\lambda_{1}$ in Eq.~\ref{l-all}. With $\lambda_{1}$=15 and $\lambda_{2}$=2 in Eq.~\ref{l-all}, EIR achieves the best top-1 kNN accuracy with 87.6\% and 61.4\% on CIFAR-10 and CIFAR-100, respectively.

%-----------------------------------------------------------------
% Table 3
\setlength{\tabcolsep}{8pt}
\begin{table}
\caption[]{Evaluation on STL-10 dataset with ResNet-18 by performing linear classifier (LC) and $k$NN ($k$=200) top-1 accuracy. Performance of other methods is copied from PSLR~\cite{Ye_2020_PSLR} and aISIF~\cite{Ye_2020_Embedding}.}
\begin{center}

\begin{tabular}{l|c|c|c|c}
\hline
\noalign{\smallskip}
Methods &Pre-train &Fine-tune &LC &kNN \\
\noalign{\smallskip}
\hline
\hline
\noalign{\smallskip}

   Random  &- &5K &28.7 &22.4  \\
   Supervised &5K &5K &83.0 &82.9 \\
\hline \noalign{\smallskip}
  
   IR &5K &5K &62.3 &66.8 \\
%   IR(NCE) &5K &5K &61.9 &66.2 \\
   DC(100) &5K &5K &56.5 &61.2 \\
   ISIF
   &5K &5K &69.5 &74.1 \\
   aISIF &5K &5K &72.2 &74.0 \\
   
   \hline \noalign{\smallskip}
   IRaug &5K &5K &69.6 &70.0 \\
    Ours  &5K &5K &\bf{76.4} &\bf{78.2} \\
   
   \hline\hline \noalign{\smallskip}
%   Exemplar~\cite{Dosovitskiy_2016_Exemplar}  &105K &75.4 &- \\
%   CPC &105K &5K &77.0 &80.8 \\
   AND &105K &5K &76.8 &80.2 \\
%   AET &105K &5K &76.6 &76.9 \\
%   AVT &105K &5K &77.2 &78.1 \\
   
   CMC &105K &5K &77.4 &81.2 \\
   ISIF &105K &5K &77.9 &81.6 \\
   PSLR &105K &5K &78.8 &83.2 \\
   aISIF &105K &5K &\bf{82.8} &83.9 \\
   
\hline \noalign{\smallskip}
   IRaug  &105k &5K &74.4 &81.0 \\
    Ours   &105K &5K &79.3 &\bf{84.7} \\
\hline
\end{tabular}

\label{tab:table3}
\end{center}
\end{table}
\setlength{\tabcolsep}{1.4pt}

%-------------------
%--------------------------------------
%Table 4
\setlength{\tabcolsep}{20pt}
\begin{table}
\caption{Evaluation on ImageNet-100 dataset with ResNet-18 by performing linear classifier (LC) and $k$NN ($k$=200) top-1 accuracy.}
\begin{center}

\begin{tabular}{l|cc}
\hline

\noalign{\smallskip}
Methods  &LC &kNN\\

\hline \hline
\noalign{\smallskip}
     Random  &6.0  &11.0\\
     IR  &60.2 &59.1\\
     DC$^*$ &56.4 &51.8 \\
     ISIF   &62.5 &59.5\\
     MoCo   &64.4 &64.2\\
     \hline \noalign{\smallskip}
     IRaug  &61.5 &61.5 \\
     
     Ours  &\bf{66.4} &\bf{66.5}\\
\hline
      
\end{tabular}

\label{tab:table4}
\end{center}
\end{table}
\setlength{\tabcolsep}{1.4pt}

%-------------------
%--------------------------------------
%Table 4
% \setlength{\tabcolsep}{3pt}
% \begin{table}
% \caption{Evaluation on ImageNet dataset subset with Resnet-18 by performing linear classifier (LC) and $k$NN(k=200) top-1 accuracy(\%).}
% \begin{center}

% \begin{tabular}{l|cc|cc|cc}
% \hline
% \noalign{\smallskip}
%  Dataset &\multicolumn{2}{c|}{ImageNet-10} &\multicolumn{2}{c|}{ImageNet-50} &\multicolumn{2}{c}{ImageNet-100}\\
%  \hline
% \noalign{\smallskip}
% Methods  &LC &kNN &LC &kNN &LC &kNN\\

% \hline \hline
% \noalign{\smallskip}
%      Random &34.2 &37.6 &10.0 &15.9 &6.0  &11.0\\
%      IR &83.4 &79.2 &70.9 &69.7 &60.2 &59.1\\
%      DC$^*$ &84.6 &82.8 &67.9 &65.7 &56.4 &51.8 \\
%      ISIF  &87.4 &86.6 &74.6 &72.6 &62.5 &59.5\\
%      MoCo  &84.0 &81.4 &72.8 &72.6 &64.4 &64.2\\
%      \hline \noalign{\smallskip}
%      IRaug &84.8 &83.6 &72.3 &71.3 &61.5 &61.5 \\
     
%      Ours  &\bf{90.8} &\bf{90.4} &\bf{76.5} &\bf{78.2} &\bf{66.4} &\bf{66.5}\\
% \hline
      
% \end{tabular}

% \label{tab:table4}
% \end{center}
% \end{table}
% \setlength{\tabcolsep}{1.4pt}

\subsection{Image Classification}
After fully training networks, we then evaluate the quality of the learned visual representations and feature embedding on image classification. The backbone is ResNet18 and the classification is evaluated by kNN and linear classification. The kNN classification uses feature embedding directly. Linear classification (LC) needs to train a fully connected linear classifier which is initialized from scratch, and the convolution layers are frozen. 
Considering the difference of image resolution for different datasets, we evaluate the performance on datasets from low resolution to high resolution.

%------------------------
% Table 5
\setlength{\tabcolsep}{2.5pt}
\begin{table}
\caption{Comparison of retrieval performance on CIFAR-10, CIFAR-100, CUB-200 and Car-196. The performance of other methods is reimplemented with their released codes. We implement DC with contrastive loss under 1000 clusters, AND with 2 rounds under 1 neighbour.}
\begin{center}

\begin{tabular}{l|cccc|cccc}
\hline\noalign{\smallskip}
 Acc(\%) &R@1 &R@2 &R@4 &R@8   &R@1 &R@2 &R@4 &R@8   \\
 \hline
\hline \noalign{\smallskip}
 Dataset &\multicolumn{4}{c|}{CIFAR-10} &\multicolumn{4}{c}{CIFAR-100} \\
 \hline
% \noalign{\smallskip}

\hline

\noalign{\smallskip}
% \hline
    Random  &25.0  &40.1 &58.2 &75.8  &8.2 &13.1 &19.8 &29.4   \\
     IR &70.8 &82.5 &90.0 &95.4 &36.0 &46.3 &56.9 &68.0   \\
     DC &75.2 &83.5 &89.2 &93.4  &36.3 &46.3 &56.9 &66.8   \\
     ISIF  &74.7 &84.6 &91.3 &95.8  &40.8 &51.4 &62.1 &72.5   \\
     AND &77.4 &85.8 &92.2 &96.3  &41.8 &52.0 &62.8 &72.8   \\
    %  MoCo &56.1 &70.4 &82.8 &91.0 &28.7 &37.5 &47.3 &58.0   \\
    %  PCP &77.5 &86.3 &92.3 &96.2  &43.9 &54.1 &64.2 &73.2   \\
     \hline\noalign{\smallskip}
     
     IRaug &73.3 &83.9 &91.4 &96.1  &39.7 &50.1 &60.5 &71.5  \\
     
     Ours   &\bf{79.9} &\bf{88.5} &\bf{93.6} &\bf{97.1}  &\bf{45.6} &\bf{56.5} &\bf{66.9} &\bf{76.5}  \\
    
    \hline\hline\noalign{\smallskip}
 Dataset  &\multicolumn{4}{c|}{CUB-200} &\multicolumn{4}{c}{Car-196} \\
 \hline
% \noalign{\smallskip}
% Methods  &R@1 &R@2 &R@4 &R@8  &R@1 &R@2 &R@4 &R@8 \\

\hline

\noalign{\smallskip}
% \hline
    Random  &3.4  &6.1 &10.4 &18.1 &4.5 &7.1 &11.1 &17.5  \\
     IR &8.0 &12.7 &20.1 &30.3 &19.9 &27.6 &37.4 &49.4  \\
     DC$^*$  &7.9 &12.7 &20.5 &30.8  &15.7 &22.1 &31.6 &43.4  \\
     ISIF  &12.6 &19.2 &28.3 &39.3  &26.6 &36.0 &46.5 &58.4  \\
     AND &11.9 &18.8 &26.8 &38.7  &22.6 &30.8 &41.3 &53.7  \\
    %  MoCo &9.1 &14.1 &21.1 &30.3  &18.7 &25.1 &33.4 &43.7  \\
    %  PCP &12.8 &19.6 &28.5 &40.4  &\bf{28.6} &\bf{38.6} &\bf{49.5} &61.0  \\
     \hline\noalign{\smallskip}
     
     IRaug &11.1 &17.1 &26.4 &37.2 &23.1 &31.1 &41.7 &53.9  \\
     
     Ours  &\bf{14.6} &\bf{22.1} &\bf{32.6} &\bf{44.1}  &\bf{28.5} &\bf{38.1} &\bf{49.5} &\bf{62.1}  \\

\hline

\end{tabular}

\label{tab:table5}
\end{center}
\end{table}
\setlength{\tabcolsep}{1.4pt}

As shown in Table~\ref{tab:table2}, on CIFAR-10 dataset under 1-round training with $k$ equal to 5/20/200, our proposed method consistently achieves the best performance. After 5-rounds training, our method achieves 89.2\% for $k=20$, which surpasses all other published methods. Moreover, the performance of our method with only 2-rounds training outperforms those of other methods with 5-rounds training. Because the performance of SimCLR~\cite{Chen_2020_SimCLR} relies on large batch size, MLP head and stronger data augmentation, it is infeasible to compare EIR with SimCLR under the similar experiment settings. The comparison of performance for different backbone is shown in the supplementary material. 

We then conduct experiments on STL-10 dataset, and report LC results with $conv5$ layer and kNN top-1 results with $k=200$ on 5K labeled set under 5K and 105K pre-training respectively.
As shown in Table~\ref{tab:table3}, our method achieves 76,4\% accuracy with LC and 78.2\%  accuracy with kNN, which outperforms state-of-the-art aISIF~\cite{Ye_2020_Embedding} by 4.2\% on 5K pre-train setting. For 105K pre-train setting, our method achieves 84.7\% accuracy with kNN, which outperforms aISIF by 0.8\%. With LC, aISIF is superior to our method probably because our method needs more training iterations for convergence.

For ImageNet-100 evaluation, we report the LC results on $conv5$ layer and kNN top-1 results with $k=200$.
As shown in Table~\ref{tab:table4}, our method achieves 66.4\% and 66.5\% accuracy for LC and KNN respectively, which outperforms MoCo at least 2\% accuracy and surpasses other methods with large margins.

\subsection{Image Retrieval}
Image retrieval experiments are conducted to evaluate the discriminative ability of the learned feature embedding. 
% We evaluate the retrieval performance on dataset which test set with seen(CIFAR-10 and CIFAR-100) and fine-grained unseen(CUB-200 and Car-196) categories. We report the retrieval performance by top-1, top-2, top-4 and top-8 ranking list.
%%
As shown in Table~\ref{tab:table5}, EIR achieves 79.9\%, 88.5\%, 93.6\% and 97.1\% accuracy for R@1, R@2, R@4 and R@8 on CIFAR-10 dataset, and 45.6\%, 56.5\%, 66.9\% and 76.5\% accuracy on CIFAR-100 dataset, which shows the best performance on these two seen datasets. 
%And we can find that the performance is decreasing with the increasing of categories, there are about 20\%~30\% descent from CIFAR-10 to CIFAR-100.%

%%
For more challenging fine-grained unseen dataset, our method achieves 14.6\%, 22.1\%, 32.6\% and 44.1\% accuracy on CUB-200 dataset and 28.5\%, 38.1\%, 49.5\% and 62.1\% accuracy on Car-196 dataset, which achieves stat-of-the-art performance. The retrieval performance on Car-196 surpasses the performance in CUB-200 with large margin, which may result from that birds are more hard to discriminate than cars.

%%%%%%%%%%%%%%%%%%%%%%%%%%%%%%%%%%%%%%%%%%%%%%%%%%%%%%%%
\begin{figure}
\centering
% width=1.0\columnwidth
\includegraphics[width=0.95\columnwidth]{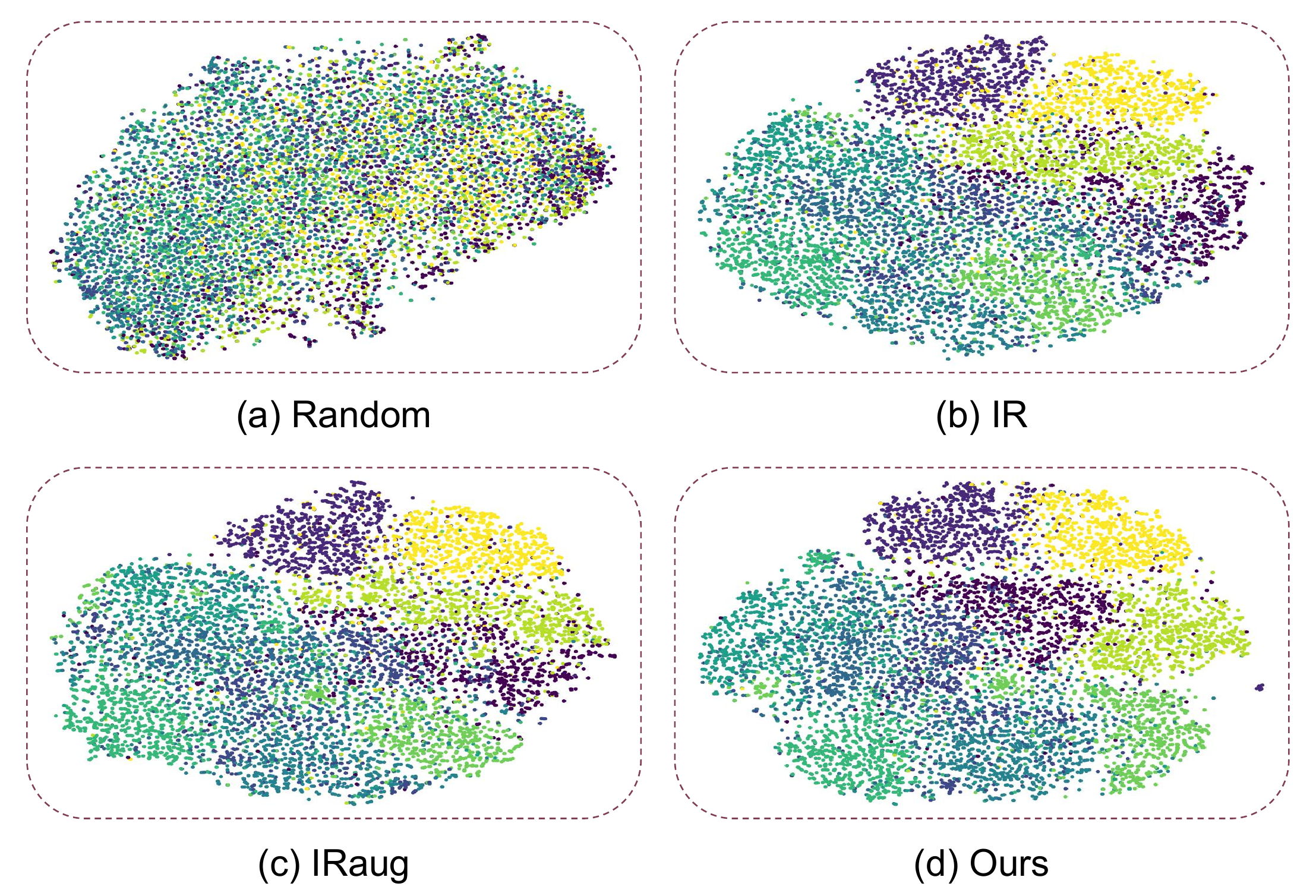}
\caption[]{Visualization of embedding space by t-SNE. (Best viewed in color)
}
\label{fig:fig5}
\end{figure}

\subsection{Visualization}
%%
% In this subsection, we analyze the feature embedding via visualizations. 
To show the effectiveness of positive pairs attracting while negative pairs repelling, we visualize the embedding features on test set of CIFAR-10 by embedding the  feature embedding into 2-dimensional space with t-SNE~\cite{tsne}. Figure~\ref{fig:fig5} shows that instance discrimination based methods achieve the goal that positive attracting while negative pairs repelling, and our method is more discriminative.

%%%%%%%%%%%%%%%%%%%%%%%%%%%%%%%%%%%%%%%%%%%%%%%%%%%%%%%%
\begin{figure}
\centering
% width=1.0\columnwidth
\includegraphics[width=0.95\columnwidth]{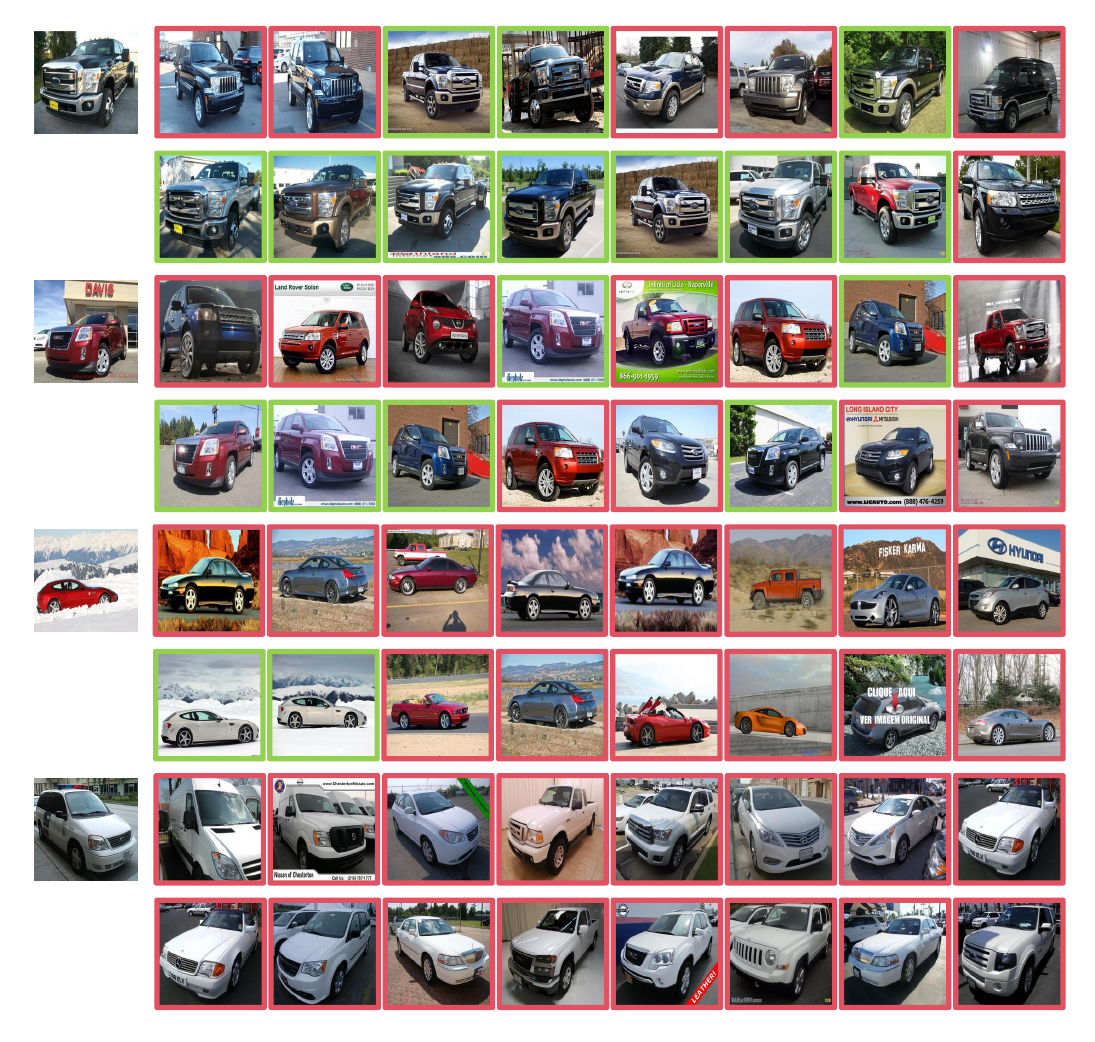}
\caption[]{ The 8 nearest neighbourhood retrieval results by baseline method IRaug (top) and EIR (down). The positive retrieved results are framed in green while negative in pink.
}
\label{fig:fig6}
\end{figure}

To help qualitatively illustrate the effectiveness of our proposed method, Figure~\ref{fig:fig6} shows the 8 nearest neighbourhood retrieval results on Car-196 test set including success and failure cases. The first column is 4 query images, and each one follows 2 rows retrieval results including IRaug (top) and EIR (down), and each row is ordered by cosine similarity. The 8 nearest neighbours of the 4-th query image are all false positive, they share high similarity on appearance, color and posture. However, from the results of first 3 query images, we can find that our model can discover positive samples with different colors, which shows that the features learned 
by our method focus on high-level semantic information.

\section{Conclusion}

In this paper, we present an unsupervised feature embedding method by exploring instance relations (EIR), which includes intra-instance multi-view relation and inter-instance interpolation relation.
We explore intra-instance multi-view relation by aligning the distance distribution between two augmented samples and each sample in the training set.
We then explore inter-instance interpolation relation by transferring the ratio of information for image sample interpolation from pixel space to feature space.
Our experimental results show that our method achieves state-of-the-art or comparable performance on image classification and retrieval. In the future, more attention will be paid to relate the different instances of the same semantic category for unsupervised learning.

\section*{Acknowledgement}

This work is supported by the National Key R\&D Program of China (2017YFB1002400), the Open Research Project of the State Key Laboratory of Media Convergence and Communication, Communication University of China, China (No. SKLMCC2020KF004), the Beijing Municipal Science \& Technology Commission (Z191100007119002), the Key Research Program of Frontier Sciences, CAS, Grant NO ZDBS-LY-7024, the National Natural Science Foundation of China (No. 62006221).

\bibliographystyle{named}
\bibliography{ijcai21}

\begin{thebibliography}{}

\bibitem[\protect\citeauthoryear{Caron \bgroup \em et al.\egroup
  }{2018}]{Caron_2018_DC}
Mathilde Caron, Piotr Bojanowski, Armand Joulin, and Matthijs Douze.
\newblock Deep clustering for unsupervised learning of visual features.
\newblock In {\em ECCV}, pages 132--149, 2018.

\bibitem[\protect\citeauthoryear{Chen \bgroup \em et al.\egroup
  }{2020}]{Chen_2020_SimCLR}
Ting Chen, Simon Kornblith, Mohammad Norouzi, and Geoffrey Hinton.
\newblock A simple framework for contrastive learning of visual
  representations.
\newblock In {\em ICML}, pages 1597--1607, 2020.

\bibitem[\protect\citeauthoryear{Coates \bgroup \em et al.\egroup }{2011}]{STL}
Adam Coates, Andrew Ng, and Honglak Lee.
\newblock An analysis of single-layer networks in unsupervised feature
  learning.
\newblock In {\em AISTATS}, pages 215--223, 2011.

\bibitem[\protect\citeauthoryear{Doersch \bgroup \em et al.\egroup
  }{2015}]{Doersch_2015_Context}
Carl Doersch, Abhinav Gupta, and Alexei~A. Efros.
\newblock Unsupervised visual representation learning by context prediction.
\newblock In {\em ICCV}, pages 1422--1430, 2015.

\bibitem[\protect\citeauthoryear{Gidaris \bgroup \em et al.\egroup
  }{2018}]{Gidaris_2018_Rotnet}
Spyros Gidaris, Praveer Singh, and Nikos Komodakis.
\newblock Unsupervised representation learning by predicting image rotations.
\newblock In {\em ICLR}, 2018.

\bibitem[\protect\citeauthoryear{He \bgroup \em et al.\egroup }{2016}]{ResNet}
Kaiming He, Xiangyu Zhang, Shaoqing Ren, and Jian. Sun.
\newblock Deep residual learning for image recognition.
\newblock In {\em CVPR}, pages 770--778, 2016.

\bibitem[\protect\citeauthoryear{He \bgroup \em et al.\egroup
  }{2020}]{He_2020_Moco}
Kaiming He, Haoqi Fan, Yuxin Wu, Saining Xie, and Ross Girshick.
\newblock Momentum contrast for unsupervised visual representation learning.
\newblock In {\em CVPR}, pages 9729--9738, 2020.

\bibitem[\protect\citeauthoryear{Huang \bgroup \em et al.\egroup
  }{2019}]{Huang_2019_AND}
Jiabo Huang, Qi~Dong, Shaogang Gong, and Xiatian Zhu.
\newblock Unsupervised deep learning by neighbourhood discovery.
\newblock In {\em ICML}, pages 2849--2858, 2019.

\bibitem[\protect\citeauthoryear{Iscen \bgroup \em et al.\egroup
  }{2018}]{mom_2018_metric}
Ahmet Iscen, Giorgos Tolias, Yannis Avrithis, and Ondrej Chum.
\newblock Mining on manifolds: Metric learning without labels.
\newblock In {\em CVPR}, pages 7642--7651, 2018.

\bibitem[\protect\citeauthoryear{Krause \bgroup \em et al.\egroup }{2013}]{Car}
Jonathan Krause, Michael Stark, Jia Deng, and Fei-Fei. Li.
\newblock 3d object representations for fine-grained categorization.
\newblock In {\em ICCVW}, pages 554--561, 2013.

\bibitem[\protect\citeauthoryear{Krizhevsky and Hinton}{2009}]{CIFAR}
Alex Krizhevsky and Geoffrey~E. Hinton.
\newblock Learning multiple layers of features from tiny images.
\newblock In {\em Citeseer}, 2009.

\bibitem[\protect\citeauthoryear{Kullback}{1959}]{KL}
Solomon Kullback.
\newblock {\em Information theory and statistics}.
\newblock Wiley, 1959.

\bibitem[\protect\citeauthoryear{Larsson \bgroup \em et al.\egroup
  }{2016}]{Guatav_2016_Color}
Gustav Larsson, Michael Maire, and Gregory Shakhnarovich.
\newblock Learning representations for automatic colorization.
\newblock In {\em ECCV}, pages 577--593, 2016.

\bibitem[\protect\citeauthoryear{Noroozi and
  Favaro}{2016}]{Noroozi_2016_Jigsaw}
Mehdi Noroozi and Paolo Favaro.
\newblock Unsupervised learning of visual representations by solving jigsaw
  puzzles.
\newblock In {\em ECCV}, pages 69--84, 2016.

\bibitem[\protect\citeauthoryear{Russakovsky \bgroup \em et al.\egroup
  }{2015}]{ImageNet-IJCV15}
Olga Russakovsky, Jia Deng, Hao Su, Jonathan Krause, Sanjeev Satheesh, Sean Ma,
  Zhiheng Huang, Andrej Karpathy, Aditya Khosla, Michael Bernstein,
  Alexander~C. Berg, and Fei-Fei Li.
\newblock Imagenet large scale visual recognition challenge.
\newblock {\em IJCV}, 115(3):211--252, 2015.

\bibitem[\protect\citeauthoryear{Tian \bgroup \em et al.\egroup
  }{2020}]{Tian_2019_CMC}
Yonglong Tian, Dilip Krishnan, and Phillip Isola.
\newblock Contrastive multiview coding.
\newblock In {\em ECCV}, pages 776--794, 2020.

\bibitem[\protect\citeauthoryear{Van~der Maaten and Hinton}{2008}]{tsne}
Laurens Van~der Maaten and Geoffrey Hinton.
\newblock Visualizing data using t-sne.
\newblock {\em JMLR}, 9(11):2579--2605, 2008.

\bibitem[\protect\citeauthoryear{Wah \bgroup \em et al.\egroup }{2011}]{CUB}
Catherine Wah, Steve Branson, Peter Welinder, Pietro Perona, and Serge
  Belongie.
\newblock The caltech-ucsd birds-200-2011 dataset, 2011.

\bibitem[\protect\citeauthoryear{Wu \bgroup \em et al.\egroup
  }{2018}]{Wu_2018_IR}
Zhirong Wu, Yuanjun Xiong, Stella~X. Yu, and Dahua Lin.
\newblock Unsupervised feature learning via non-parametric instance
  discrimination.
\newblock In {\em CVPR}, pages 3733--3742, 2018.

\bibitem[\protect\citeauthoryear{Yang \bgroup \em et al.\egroup
  }{2016}]{Yang_2016_CVPR}
Jianwei Yang, Devi Parikh, and Dhruv Batra.
\newblock Joint unsupervised learning of deep representations and image
  clusters.
\newblock In {\em CVPR}, pages 5147--5156, 2016.

\bibitem[\protect\citeauthoryear{Ye and Shen}{2020}]{Ye_2020_PSLR}
Mang Ye and Jianbing Shen.
\newblock Probabilistic structural latent representation for unsupervised
  embedding.
\newblock In {\em CVPR}, pages 5457--5466, 2020.

\bibitem[\protect\citeauthoryear{Ye \bgroup \em et al.\egroup
  }{2019}]{Ye_2019_IS}
Mang Ye, Xu~Zhang, Pong~C. Yuen, and Shih-Fu Chang.
\newblock Unsupervised embedding learning via invariant and spreading instance
  feature.
\newblock In {\em CVPR}, pages 6210--6219, 2019.

\bibitem[\protect\citeauthoryear{Ye \bgroup \em et al.\egroup
  }{2020}]{Ye_2020_Embedding}
Mang Ye, Jianbing Shen, Xu~Zhang, Pong~C Yuen, and Shih-Fu Chang.
\newblock Augmentation invariant and instance spreading feature for softmax
  embedding.
\newblock {\em TPAMI}, 2020.

\bibitem[\protect\citeauthoryear{Yun \bgroup \em et al.\egroup
  }{2019}]{cutmix_2019_iccv}
Sangdoo Yun, Dongyoon Han, Sanghyuk Chun, Seong~Joon Oh, Youngjoon Yoo, and
  Junsuk Choe.
\newblock Cutmix: Regularization strategy to train strong classifiers with
  localizable features.
\newblock In {\em ICCV}, pages 6022--6031, 2019.

\bibitem[\protect\citeauthoryear{Zhan \bgroup \em et al.\egroup
  }{2020}]{Zhan_2020_ODC}
Xiaohang Zhan, Jiahao Xie, Ziwei Liu, Yew~Soon Ong, and Chen~Change Loy.
\newblock Online deep clustering for unsupervised representation learning.
\newblock In {\em CVPR}, pages 6688--6697, 2020.

\bibitem[\protect\citeauthoryear{Zhang \bgroup \em et al.\egroup
  }{2017}]{Zhang_2017_splitbrain}
Richard Zhang, Phillip Isola, and Alexei~A. Efros.
\newblock Split-brain autoencoders: Unsupervised learning by cross-channel
  prediction.
\newblock In {\em CVPR}, pages 1058--1067, 2017.

\bibitem[\protect\citeauthoryear{Zhang \bgroup \em et al.\egroup
  }{2018}]{mixup_2018_iclr}
Hongyi Zhang, Moustapha Cissé, Yann~N. Dauphin, and David Lopez-Paz.
\newblock mixup: Beyond empirical risk minimization.
\newblock In {\em ICLR}, 2018.

\bibitem[\protect\citeauthoryear{Zhuang \bgroup \em et al.\egroup
  }{2019}]{Zhuang_2019_LA}
Chengxu Zhuang, Alex~Lin Zhai, and Daniel Yamins.
\newblock Local aggregation for unsupervised learning of visual embeddings.
\newblock In {\em ICCV}, pages 6002--6012, 2019.

\end{thebibliography}

\end{document}